%% file: main.tex
\documentclass[conference]{IEEEtran}
\usepackage{times}

\usepackage[numbers]{natbib}
\usepackage{multicol}
\usepackage[bookmarks=true]{hyperref}
\usepackage{graphicx}
\usepackage{wrapfig}
\usepackage{caption}
\usepackage{amssymb}
\usepackage{amsmath}
\usepackage{xcolor}
\usepackage{subfigure}
\usepackage{algorithm} 
\usepackage{algpseudocode} 
\usepackage[normalem]{ulem}

\begin{document}

\title{Online Planning  for Cooperative Air-Ground Robot Systems with Unknown Fuel Requirements}

\author{Ritvik Agarwal, Behnoushsadat Hatami, Alvika Gautam and Parikshit Maini}



%

\maketitle
\input{Abstract}


\input{Introduction}
\input{Solution_Approach}
\input{exp_res_mrs}
\bibliographystyle{unsrt}
\bibliography{ormrs_references, ref_hatami}

\end{document}

%% file: Abstract.tex
\begin{abstract}

We consider an online variant of the fuel-constrained UAV routing problem with a ground-based mobile refueling station (FCURP-MRS), where targets incur unknown fuel costs. We develop a two-phase solution: an offline heuristic-based planner computes initial UAV and UGV paths, and a novel online planning algorithm that dynamically adjusts rendezvous points based on real-time fuel consumption during target processing. Preliminary Gazebo simulations demonstrate the feasibility of our approach in maintaining UAV-UGV path validity, ensuring mission completion. Link to video: \href{https://youtu.be/EmpVj-fjqNY}{https://youtu.be/EmpVj-fjqNY}

\end{abstract}

%% file: Introduction.tex
\section{Introduction}\label{sec:intro}

Cooperative air-ground robot systems leverage complementary capabilities in range, speed and sensing resolution to enhance mission efficiency. A key research focus in this area has been the use of ground robots as mobile refueling stations \cite{maini2015cooperation,maini2019cooperative,mathew2015multirobot,du2020cooperative} to extend UAV operational range in space and time. Significant literature exists for offline rendezvous planning \cite{zhang2022routeplanning} and cooperative strategies \cite{mondal2023optimizing,ramasamy2021cooperative} to address constraints such as speed differentials, terrain limitations, recharging windows \cite{mondal2024robustcollaboration} and energy constraints \cite{diller2023energy}. In practical applications such as, pick/drop of parcels, sensor data collection, and aerial imaging, UAVs must process points of interest in real time, requiring additional maneuvers, causing unpredictable fuel consumption that offline planners cannot accommodate. A recent line of work \cite{ramasamy2024optimizing,mondal2024attention} develops partially online planning strategies for UAV-UGV missions for new targets that appear during mission execution by replanning during rendezvous stops. There does not exist any work in the literature that performs fully online real-time planning for collaborative UAV-UGV missions with unknown target processing costs.

In this work, we address an online version of the Fuel-Constrained UAV Routing Problem with a ground-based Mobile Refueling Station (FCURP-MRS). UAV has limited fuel capacity. It must visit a set of targets, whose locations are known a priori, and return to start location. Target processing times and consequently the fuel consumed to process each target are not known until the processing is fully complete. UAV may rendezvous with the ground robot as needed to refuel. The two robots are assumed to have perfect real-time communication, localization and navigation capabilities. UAV travels with a constant ground speed (equal to average speed) and is assumed to have a constant rate of fuel consumption independent of the flight maneuver. As a result distance traveled by the UAV is proportional to time of flight and fuel consumed. UGV has a fixed known maximum speed and does not run out of fuel within the scope of the mission.

Due to previously unknown fuel consumption to process each target, precomputed plans become infeasible, necessitating online planning for the UAV and UGV to adjust refueling (rendezvous) site locations and paths based on real-time fuel information. The cooperative aerial-ground robot routing problem with unknown target processing times, is then defined as follows: \textit{Develop an online planner for a fuel-constrained UAV and mobile refueling UGV to ensure that the UAV visits and processes all targets, avoids fuel depletion, rendezvous as needed to refuel, and minimizes total path length.}

%% file: Solution_Approach.tex
\section{Solution Approach}\label{sec:approach}

\input{use_cases_figure}

We develop a two phase solution approach. Phase 1 involves developing an offline plan for the two robots that satisfies the problem constraints. Phase 2 involves online planning during the mission to accommodate unknown target processing costs.



\subsection{Phase 1: Offline Path Planning} The offline planner assumes that target processing is instantaneous and does not consume any fuel.  This version of the problem is NP-hard as shown in our previous work \cite{maini2019cooperative}. We have developed exact methods \cite{maini2019cooperative} and computationally efficient heuristics  \cite{maini2015cooperation,maini2019cooperative,arora2019route} for the offline problem in our previous work. We adopt the TSP based heuristic approach developed in \cite{maini2019cooperative} for offline path planning in Phase 1. The interested reader is referred to \cite{maini2019cooperative} for details of the offline heuristic algorithms. The output of the offline planner is a UAV path, $\mathbb{P}$, that consists of segments $\mathbb{P}_{1}\ldots\mathbb{P}_{K}$, where each segment starts and ends at a refueling site, visiting a set of targets in between. Thus, each path segment comprises the following structure: $\mathbb{P}_k = [d_{k-1}, t_{k1}, t_{k2}, \dots, t_{ks}, d_k]$, where $d_k$ is the $k^{th}$ refueling site on UAV's path and $[t_{k1} \dots t_{ks}]$ are the targets visited in that segment. The length of each path segment is constrained by $U$ (maximum flight range of the UAV). Let $U_k$ denote the actual length of path segment $\mathbb{P}_k$; then the available excess fuel for segment $\mathbb{P}_k$ is given by $z_k = U - U_k$ (since path length and fuel consmed are proportional). UGV path comprises the sequence of refueling sites on the UAV path. UGV’s traversal between consecutive refueling sites on its path is limited by $R$ (maximum distance the UGV can travel within UAV's flight range). 

\subsection{Phase 2: Online Replanning}
We recall that the UAV path $\mathbb{P}$ computed in Phase 1 accounts for UAVs traversals between targets and refueling sites. The sole source of uncertainty in the system during online execution is attributed to the excess fuel consumed during target processing. Let $\tau_i$ be the fuel needed to process target $t_i$. The online planner must account for this fuel consumption and ensure that the UAV never runs out of fuel and all targets are processed successfully. The proposed planning algorithm operates by maintaining the sequence in which the targets are visited on the UAV path $\mathbb{P}$ computed in Phase 1 and re-planning the refueling site locations in real-time. It maintains the following invariant throughout the mission: \textit{During any path segment $\mathbb{P}_k$, the next refueling site $d_k$ on the UAV path must be reachable by both, the UAV and the UGV, at all times}.


The online planner models each path segment, $\mathbb{P}_k$, as a \textit{string} tied to the UAV at one end and the refueling site $d_k$ on the other, running along $\mathbb{P}_k$. The string length represents the current remaining fuel in the UAV. As the UAV consumes fuel, the string shortens by a proportional length. At the start of a segment, the string length equals $U$. We recall from Phase 1, that $z_k$ is the amount of excess fuel available in a path segment of the UAV. Hence, if $z_k > 0$, the string has some \textit{slack} in it and does not have any tension. During target processing, the UAV starts to use up the excess fuel and $z_k$ starts to reduce. Once $z_k = 0$, all slack is consumed, and the string becomes tensed. At this stage, as the UAV consumes more fuel during target processing, the string keeps shortening and the refueling site starts backtracking along $\mathbb{P}_k$ towards the UAV. This real-time backtracking ensures that the refueling site remains within the UAV’s reach. The UAV shares the updated refueling site location with the UGV in real-time, allowing the UGV to adjust its path and rendezvous efficiently. The UGV's onboard navigation controller navigates it to the new location. It may be noted, that backtracking only occurs during target processing, since fuel required for UAV traversal is already accounted for in the offline path planning. The backtracking also checks to ensure that the refueling site remains reachable by the UGV. If the backtracking results in refueling site going out of range of the UGV, rest of the path segment (including the current target) is abandoned and the online planner initiates a rendezvous sequence. The algorithm's execution can be summarized in the following use-cases:
\subsubsection{Case 1 - $z_k \geq \sum_{i:t_i\in\mathbb{P}_k} \tau_i$} The slack fuel $z_k$ is greater than the total fuel needed to process the targets and the UAV is able to successfully execute $\mathbb{P}_k$ and reach $d_k$ without needing to replan the refueling site (Figure \ref{fig:outline}(a)).

\subsubsection{Case 2 - Refueling site is replanned and all targets in $\mathbb{P}_k$ are visited} In this scenario, $z_k < \sum_{i:t_i\in\mathbb{P}_k} \tau_i$. The refueling site backtracks along $\mathbb{P}_k$ (as shown in Figure \ref{fig:outline}(b)) but the UAV is able to successfully process all the targets in $\mathbb{P}_k$.

\subsubsection{Case 3 - Refueling site is replanned and all targets in $\mathbb{P}_k$ cannot be visited}  In this scenario, $z_k < \sum_{i:t_i\in\mathbb{P}_k} \tau_i$. The UAV does not have enough fuel to process all targets in $\mathbb{P}_k$. As the fuel depletes, the refueling site backtracks skipping targets in $\mathbb{P}_k$ resulting in unvisited targets (see Figure \ref{fig:outline}(c)). Any unvisited targets are added on to path segment $\mathbb{P}_{k+1}$. 

\subsubsection{Case 4 - UAV adandons the target during processing}  (Figure \ref{fig:outline}(d)) When the planner computes during target processing that backtracking $d_k$ any further to process the target will result in refueling site becoming unreachable from the UGV, UAV abandons the target processing. UAV and UGV start to move towards the refueling site to rendezvous. Any unprocessed targets in $\mathbb{P}_k$ are prefixed to $\mathbb{P}_{k+1}$ in sequence.

\subsubsection{Case 5 - $U < U_k$} $\mathbb{P}_k$ is infeasible as the path needs more fuel than available. This can occur if targets were transferred to current segment from the previous segment. To fix this, $d_k$ is backtracked until $U = U_k$.






%% file: use_cases_figure.tex
\begin{figure*}[!h]
    \centering
    \includegraphics[width=0.15\linewidth]{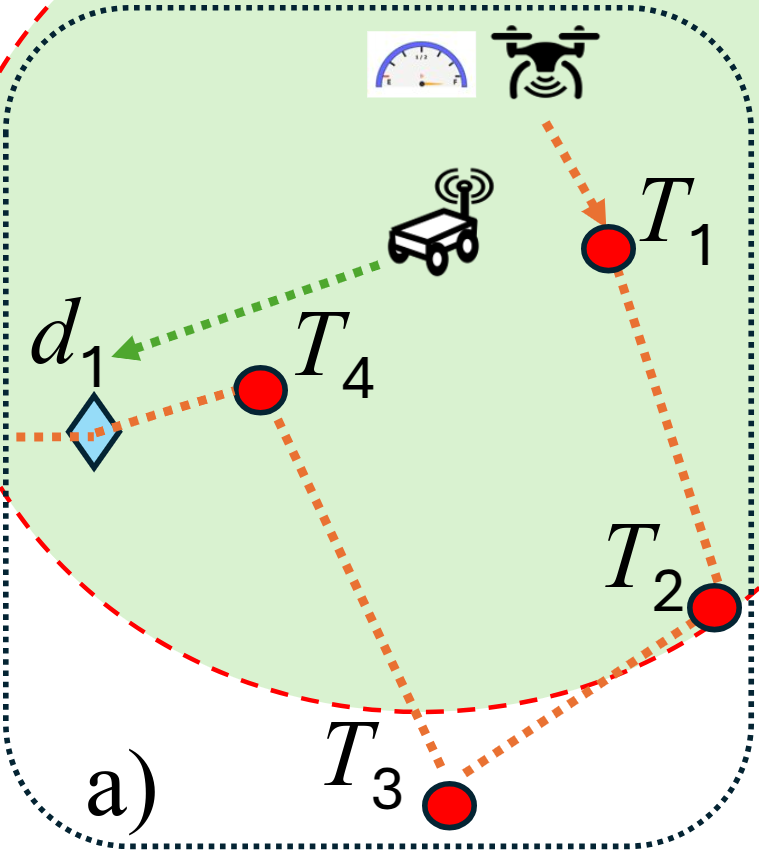}
    \includegraphics[width=0.16\linewidth]{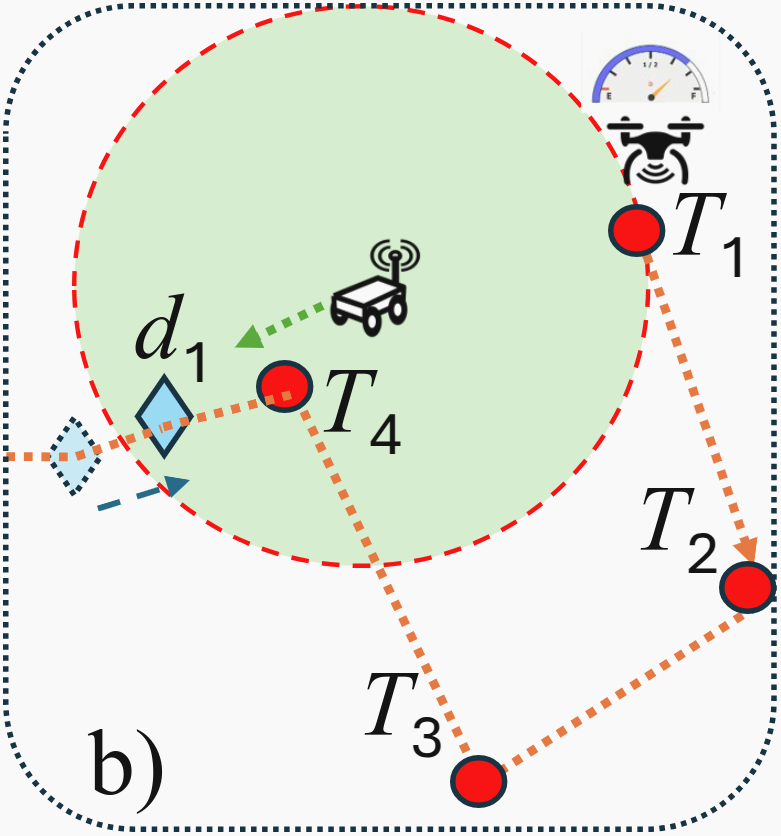}
    \includegraphics[width=0.155\linewidth]{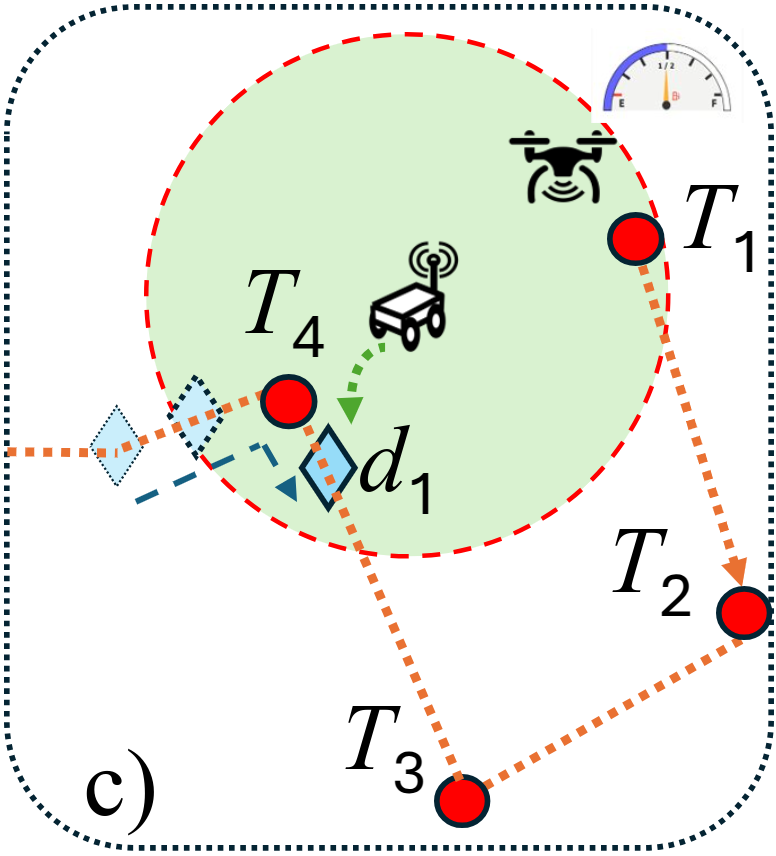}
    \includegraphics[width=0.15\linewidth]{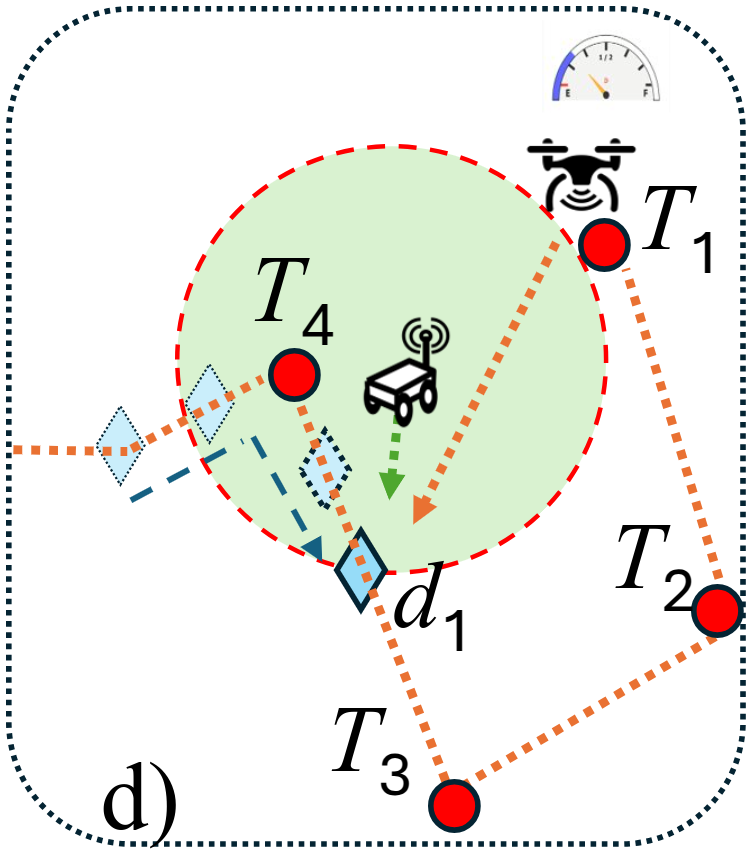}
    \includegraphics[height=0.125\textheight,width=0.34\linewidth]{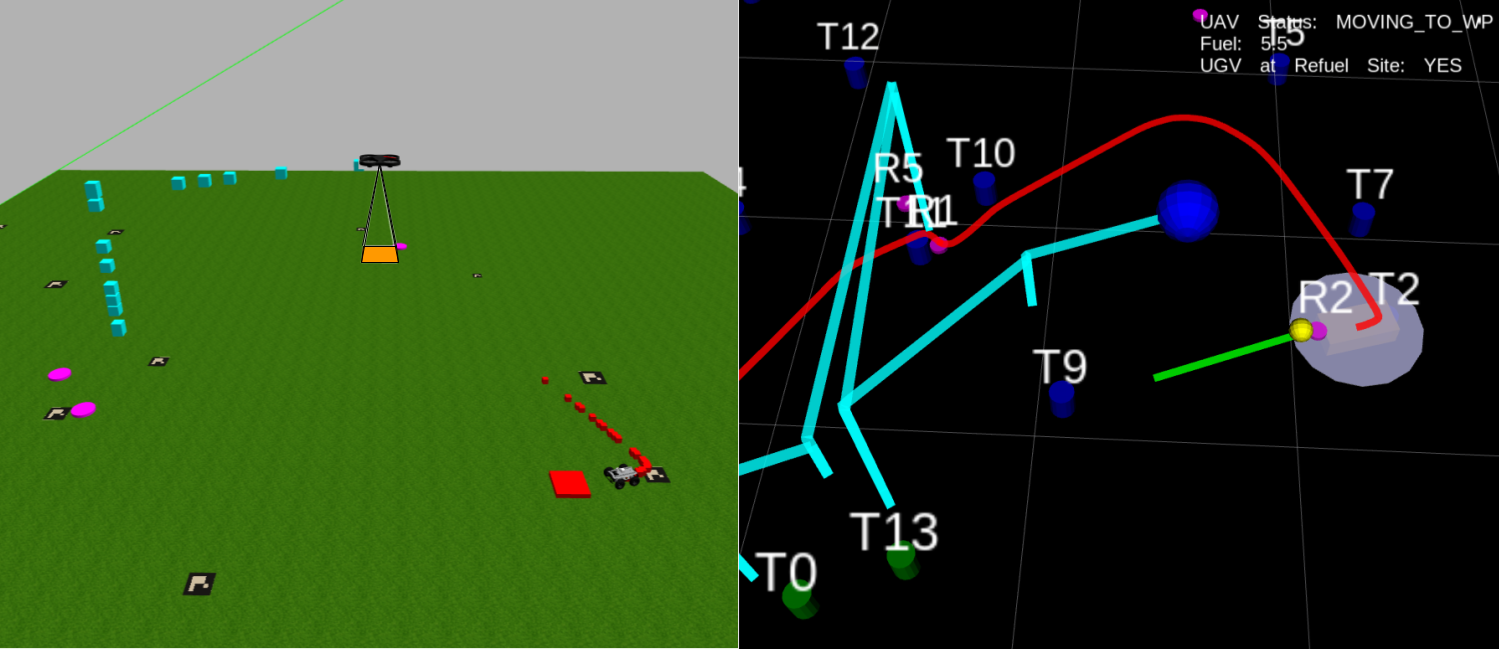}
    \captionsetup{font=small}
    \caption{Application environment with targets $T_1 \ldots T_4$ shown in red disks, a refueling site $d_1$ shown as a blue colored rhombus, a UAV, and a UGV with their paths shown with orange and green dashed lines respectively. The green circled area ahows reachable space by the UGV. (a) UAV starts a new path segment at full fuel capacity. (b) As the UAV processes $T_1$ the refueling site $d_1$ backtracks along the UAV path while remaining in the UGV’s range (case 2). (c) Extended processing at $T_1$ results in $d_1$ to backtrack and skip over $T4$ to maintain feasibility (case 3). (d) The UAV and UGV feasibility are in conflict with each other i.e. $d_1$ needs to backtrack further to maintain UAV feasibility while further backtracking will make it unreachable for the UGV. This results in UAV to abandon $T_1$ and rendezvous with UGV to refuel (case 4). (e) Gazebo-based simulation environment for online UAV-UGV planning and coordination. }
    \label{fig:outline}
\end{figure*}

%% file: exp_res_mrs.tex
\section{Simulations and Results}

The proposed approach has been implemented in ROS2 using Gazebo on a 64-bit Ubuntu 22.04 system with an Intel i7 CPU and 30 GB RAM. The UAV is modeled as a VTOL flying at $2$ $m/s$ with a cruise altitude of 10 m and a fuel capacity of 50 units (1 unit/m). The UGV is a differential-drive robot (Scout 2.0) with velocity limits of $1$ $m/s$ (linear) and 0.4 $rad/s$ (angular). Simulations are run in a flat 50 x 50 m terrain and targets are simulated using ArUco tags of varying resolution that the UAV must process at same image scale resulting in uncertain processing times(cost), thus needing real-time replanning.

Figure \ref{fig:outline}(e) shows an example mission where the UAV visits multiple targets and successfully rendezvous with the UGV at a backtracked refueling site. Future large-scale simulations will explore varying target counts (10–100), fuel capacities (50–500 units), and UAV-to-UGV speed ratios (0.2 – 1.0), while evaluating metrics such as mission length, replanning frequency, and target reassignments.

%% file: main.bbl
\begin{thebibliography}{10}

\bibitem{maini2015cooperation}
Parikshit Maini and PB~Sujit.
\newblock On cooperation between a fuel constrained uav and a refueling ugv for large scale mapping applications.
\newblock In {\em International Conference on Unmanned Aircraft Systems (ICUAS)}. IEEE, 2015.

\bibitem{maini2019cooperative}
Parikshit Maini, Kaarthik Sundar, Mandeep Singh, Sivakumar Rathinam, and PB~Sujit.
\newblock Cooperative aerial--ground vehicle route planning with fuel constraints for coverage applications.
\newblock {\em IEEE Transactions on Aerospace and Electronic Systems}, 55(6):3016--3028, 2019.

\bibitem{mathew2015multirobot}
Neil Mathew, Stephen~L Smith, and Steven~L Waslander.
\newblock Multirobot rendezvous planning for recharging in persistent tasks.
\newblock {\em IEEE Transactions on Robotics}, 31(1):128--142, 2015.

\bibitem{du2020cooperative}
Bin Du, Dengfeng Sun, Satyanarayana~Gupta Manyam, and David~W Casbeer.
\newblock Cooperative air-ground vehicle routing using chance-constrained optimization.
\newblock In {\em American Control Conference (ACC)}, pages 392--397. IEEE, 2020.

\bibitem{zhang2022routeplanning}
Mingjia Zhang, Huawei Liang, and PengFei Zhou.
\newblock Cooperative route planning for fuel-constrained ugv-uav exploration.
\newblock In {\em IEEE International Conference on Unmanned Systems (ICUS)}, pages 1047--1052, 2022.

\bibitem{mondal2023optimizing}
Md~Safwan Mondal, Subramanian Ramasamy, James~D Humann, Jean-Paul~F Reddinger, James~M Dotterweich, Marshal~A Childers, and Pranav Bhounsule.
\newblock Optimizing fuel-constrained uav-ugv routes for large scale coverage: Bilevel planning in heterogeneous multi-agent systems.
\newblock In {\em International Symposium on Multi-Robot and Multi-Agent Systems (MRS)}, pages 114--120. IEEE, 2023.

\bibitem{ramasamy2021cooperative}
Subramanian Ramasamy, Jean-Paul~F Reddinger, James~M Dotterweich, Marshal~A Childers, and Pranav~A Bhounsule.
\newblock Cooperative route planning of multiple fuel-constrained unmanned aerial vehicles with recharging on an unmanned ground vehicle.
\newblock In {\em International Conference on Unmanned Aircraft Systems (ICUAS)}, pages 155--164. IEEE, 2021.

\bibitem{mondal2024robustcollaboration}
Md~Safwan Mondal, Subramanian Ramasamy, James~D. Humann, James~M. Dotterweich, Jean-Paul~F. Reddinger, Marshal~A. Childers, and Pranav Bhounsule.
\newblock A robust uav-ugv collaborative framework for persistent surveillance in disaster management applications, 2024.

\bibitem{diller2023energy}
Jonathan Diller and Qi~Han.
\newblock Energy-aware drone path finding with a fixed-trajectory ground vehicle.
\newblock 2023.

\bibitem{ramasamy2024optimizing}
Subramanian Ramasamy, Md~Safwan Mondal, James~D Humann, James~M Dotterweich, Jean-Paul~F Reddinger, Marshal~A Childers, and Pranav~A Bhounsule.
\newblock Optimizing routes of heterogenous unmanned systems using supervised learning in a multi-agent framework: A computational study.
\newblock In {\em International Conference on Unmanned Aircraft Systems (ICUAS)}, pages 286--294. IEEE, 2024.

\bibitem{mondal2024attention}
Md~Safwan Mondal, Subramanian Ramasamy, James~D Humann, James~M Dotterweich, Jean-Paul~F Reddinger, Marshal~A Childers, and Pranav Bhounsule.
\newblock An attention-aware deep reinforcement learning framework for uav-ugv collaborative route planning.
\newblock In {\em IEEE/RSJ International Conference on Intelligent Robots and Systems (IROS)}, pages 13687--13694, 2024.

\bibitem{arora2019route}
Divansh Arora, Parikshit Maini, Pedro Pinacho-Davidson, and Christian Blum.
\newblock Route planning for cooperative air-ground robots with fuel constraints: an approach based on cmsa.
\newblock In {\em Proceedings of the Genetic and Evolutionary Computation Conference}, pages 207--214, 2019.

\end{thebibliography}
